# Automated Human Mind Reading Using EEG Signals for Seizure Detection

Virender Ranga

*Department of Computer Engineering,*
*National Institute of Technology, Kurukshetra, Haryana, India*

Shivam Gupta*

*Department of Computer Science and Engineering,*
*Indian Institute of Information Technology, Sonepat, Haryana, India*
*(Mentor National Institute of Technology, Kurukshetra, Haryana, India)*

Jyoti Meena

*Department of Computer Engineering,*
*National Institute of Technology, Kurukshetra, Haryana, India*

Priyansh Agrawal

*Department of Computer Engineering,*
*National Institute of Technology, Kurukshetra, Haryana, India*

**\*Corresponding Author Contact: Shivam Gupta , [shivi98g@gmail.com](mailto:shivi98g@gmail.com)**

# Automated Human Mind Reading Using EEG Signals for Seizure Detection

**ABSTRACT**

Epilepsy is one of the most occurring neurological disease globally emerged back in 4000 BC. It is affecting around 50 million people of all ages these days. The trait of this disease is recurrent seizures. In the past few decades, the treatments available for seizure control have improved a lot with the advancements in the field of medical science and technology. Electroencephalogram (EEG) is a widely used technique for monitoring the brain activity and widely popular for seizure region detection. It is performed before surgery and also to predict seizure at the time operation which is useful in neuro stimulation device. But in most of cases visual examination is done by neurologist in order to detect and classify patterns of the disease but this requires a lot of pre-domain knowledge and experience. This all in turns put a pressure on neurosurgeons and leads to time wastage and also reduce their accuracy and efficiency. There is a need of some automated systems in arena of information technology like use of neural networks in deep learning which can assist neurologists. In the present paper, a model is proposed to give an accuracy of 98.33% which can be used for development of automated systems. The developed system will significantly help neurologists in their performance.

## 1. INTRODUCTION

Epilepsy is one of the most occurring neurological disease globally. It is common in early life during childhood. This disease is one of the oldest recognized conditions back to 4000 BC with records. Today it is affecting around 50 million people of all ages. It is a chronic non-communicable disease involving brain [19, 26]. The trait of this disease is recurrent seizures which are small intervals of involuntary movement or jerking that can involve the entire body or the partial body. Sometimes seizure results in the loss of consciousness or dysfunction of bladder [5]. The main reasons for epileptic seizures are electrical imbalance & loss of oxygen to brain, or it can be due to some severe head injury, brain stroke and even can be due to genetic heredity [22]. Usually the site of such imbalance can be different parts of the brain. Epileptic seizures can vary in frequency from nominal to several per day. Also along with seizures epilepsy tends to have other psychological conditions, like anxiety and depression too. But seizures can be controlled, around 70% of the people could become seizure free with appropriate medical treatment [5]. But only the main step is timely diagnosis and choice of drugs used for curing.

In the past few decades, with the advancements in the field of medical science and technology now the treatments available for seizure control have improved a lot [12, 23]. Some of treatments presently available includes use AEDs which are anti-epilepsy medicine [7]. But in 70% of cases these drugs are not one of successful techniques especially when the effected part of brain is severe. In such cases only treatment available for seizure control is to remove the part of brain affected by epilepsy or sometimes use devices that can be inserted inside the body to balance out the electrical imbalance in brain. In some cases obeying a regulated diet pattern involving ketogenic diet has also helped in controlling the seizures. But in most of these available techniques identification of region of brain affected by epilepsy not for treatment but also to understand the severity plays a crucial role. Electroencephalogram (EEG) is a widely used technique for monitoring the brain activities [19]. EEG is a collection of brain waves. In broad there are two types of EEG, one of them is a wearable electrodes which is placed on the human scalp as shown in Figure 1. and the other one is implanted inside the skull which is called intracranial EEG (iEEG) [20]. EEG signals are sampled through these two electrodes with the help of electric circuit. These signals are amplified and digitalized. Afterwards signals are sent to computer where post-processing and analysis of signals are done. Extraction of features and detection of diseases such as memory loss, Autisms, Alzheimer and Epilepsy are checked.

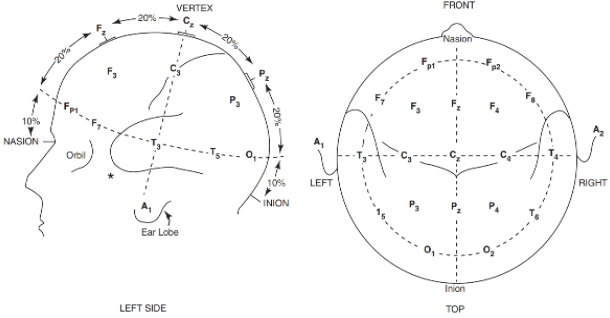

Figure 1. Wearable EEG device on head scalp

### 1.1 Diagnosis Challenges

In most of cases visual examination is done by neurologist in order to detect and classify patterns of the disease but this requires a lot of pre-domain knowledge and experience [4]. In such cases, if neurosurgeons try to seek help or advice from experts, there are issues of non-standards of recording the brain signals. Further, EEG data generated is of low quality and is difficult to differentiate EEG between healthy and patient manually [6]. This all in turns put a pressure on neurosurgeons and leads to time wastage and reducing their efficiency [9]. EEG records are also affected by noise easily. Thus there is need of some automated system using present information technology tools such as neural networks in deep learning of the EEG signals. It can assist neurologists to classifying the epileptic and non-epileptic EEG brain signals of the patients. This will enable work done by neurologists to be more precise and efficient [13].

### 1.2 Deep Neural Network

Neural network is generally set of algorithms, which works like a human brain to recognize patterns around us. It takes raw data by the help of a mechanical method such as a sensor, camera, microphone etc. The data is processed and can be labelled or unlabeled. It finds patterns in all types of real world data that is collected such as sound, text or image. A deep neural network is generally stacked neural network layers. It consists of an input layer and an output layer. Hidden layers that are added in between input and output. It defines the depth of a neural network. More hidden layers are added in the neural network to get better result since they extract those features which may not be recognized by previous layers. Basic deep neural network is shown in Figure 2.

Over the past few years a lot of neural network architectures have been proposed and have become a standard for most of the problems that has been solved using neural networks. These neural architectures are LeNet, AlexNet, VGG13, DenseNet etc. [10, 11, 14, 21].

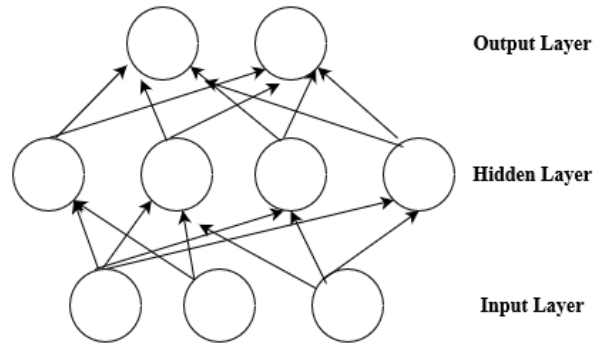

Figure 2. Basic neural network

Some of hidden layers that are used in the present study are as follows:-

- **Convolution**

Basically convolution may be imagined as mixture of information. Convolution layer can be 1D or 2D. Conv1D is used for analyzing linear data that is in the form of 1D array of data that is required to be read linearly such as text. Conv2D is mainly used when data is in matrix form such as image. In the present study, Conv1D is used since EEG signals is a single channel problem and should be modelled sequentially. Figure 3. shows an example of Conv1D.

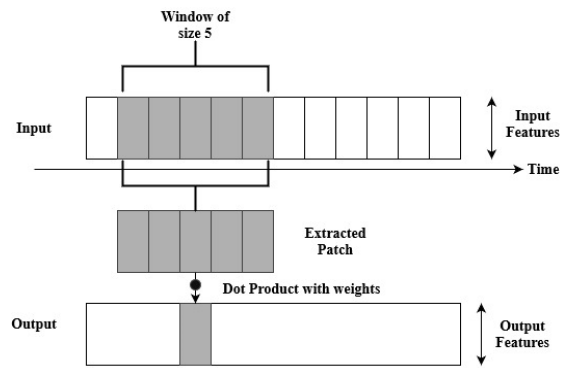

Figure 3. Convolution 1D process

- *Pooling*

  It is basically done to reduce spatial dimensions for next layer. It is either done by taking maximum value from sliding window (maximum pooling) or by doing average of all values in current window. Usually maximum pooling is used for its better performance and accuracy.

- *Activation Function*

  Activation functions are used to introduce non-linearity in the feature map. This further helps in the extraction of features. Some common activation function are given below:-
  1) Linear
  2) Sigmoid
  3) ReLU
  4) Tanh
  5) Softmax

## 2. LITERATURE REVIEW

Jian et al. (2020) exploited to form 2D matrix of electroencephalogram signals for a two dimensional convolutional neural network and achieved an accuracy of 97.82% for Task1 (Binary) [15]. Liu et al. also proposed a 2D CNN on multi bio signal that gave an accuracy of 64.5% for Task4 containing all the five classes [16]. Mao et al. (2020) came up with a combined CNN model with continuous wavelet transform for classification of epilepsy signals that gave an accuracy score of 72.49% for all the five classes [17]. Usman et al. (2020) came up with a model which extracted features using CNN and later used SVM for classification into pre-ictal and ictal with an average sensitivity and specificity of 92.7% and 90.8% respectively [25]. Nagabushanam et al. (2019) proposed a DL-LSTM for two way classification that gave an accuracy of 71.38% [18]. Acharya et al. came up with a 13 layer deep convolutional neural network that detected normal pre-ictal, and seizure classes with an accuracy, specificity, and sensitivity of 88.67%, 90.00% and 95.00%, respectively [3]. Emami et al. after applying filters converted EEG data into images using window size and achieved a binary classification accuracy of 74% for Task 1 [8].

Abedin et al. (2019) proposed an ANN that gave an accuracy of 97.33% and precision, recall, F1-score for class E as 0.96, 1 and 0.98 respectively [2].

Other than these other significant works carried out in past few years are Bhagat et al. in 2019 came up with a AGWO optimizer based deep stacked auto encoder network[4]. Abbasi et al. came up with a long short term memory (LSTM) based RNN classifier that gave an score of 94.103% for Task 1 and 94.8% for Task3 respectively [1]. Yeola et al. also designed a computer aided diagnosis (CAD) system for EEG with 11 layer deep CNN with a score of 97.38% for Task1 [27]. In 2018 Ullah et al. proposed a Pyramidal CNN model (P-CNN) that achieved a accuracy of 94.8% for Task1 [24] While reviewing the previous work, it found that in most of the work, the true nature (raw textual) of data is not preserved. It is either converted into 2-D forms, images or spectrographs to use preexisting 2D convolutional neural networks. Also the most of the work is carried out around binary classification, in which feature extraction is comparatively much easier as compare to 5 classes.

### 2.1 Dataset

Epileptic Seizure Recognition Data Set includes the samples of epilepsy patients in Department of Epileptology, Bonn University [28]. The dataset is free to use for research and education related uses. The dataset is downloaded from UCI Machine learning repository for the classification of various stages in an epileptic seizure. Samples were recorded from five patients undergoing treatment to identify cortex region so that future epilepsy can be stopped. As EEG signals are nothing but time series representation of values that depicts the neurological brain activities so the deep learning has proved to be significant helper in decoding brain activities. The signals from patients were recorded using an amplifier with 128 channels.

The recorded sample were converted into digital processed form using analog to digital convertor. The duration of each signal is about 23.6 second. Samples are broadly divided into five main categories (A-E) as shown in Table 1.

The dataset contains 11500 samples of EEG recording with each sample having length of 178 and 1 column for label category in numerical format with A-E as 5 to 1. The class A is present in Z.zip, B in O.zip, C in N.zip, D in F.zip, and E in S.zip in the original source dataset. In some works these filename are used as class names.

Table 1. Description of dataset classes

| Class Name | Remarks | Other Name |
|---|---|---|
| A | Normal Person with open eyes | healthy |
| B | Normal Person with closed eyes | healthy |
| C | Unhealthy Person Seizure free interval (hippocampal part of brain) | inter-ictal |
| D | Unhealthy Person Seizure free interval (epileptogenic part of brain) | pre-ictal |
| E | Unhealthy Person during Seizure | ictal |

## 3. RESEARCH METHODLOGIES

In the present study, complete epilepsy diagnosis process is proposed to divide into four tasks for progressive improvement in the training of the model. It is based on the dataset classifications shown in the table 1 that are significant in detection and treatment of the cortical region of brain affected by seizure.

### 3.1 TRAINING METHODOLGY

The proposed tasks are categorized as follows:-

Task 1: To identify the healthy brain activities (A, B) from brain during epilepsy seizure (E).

Task 2: To identify the healthy (A, B) wellbeing from epileptic patient (C, D, and E).

Task 3: Seizure identification between healthy (A, B), pre-ictal(C, D) and ictal (E).

Task 4: To identify between all the five states i.e. Normal with open eyes (A), Normal with closed eyes (B), non-seizure unhealthy brain C and D and during seizure E.

### 3.2 PROPOSED MODEL

The proposed model comprises mainly of a basic block formed by two layers of proposed convolutional 1D layers with a filter size of 3x3 and a padding of 1x1 along with a batch normalization layer applied after each 1D convolution operation. After the output is obtained from basic block then to prevent gradient problem, the input to basic block is added back to output of basic block.

Now the complete proposed 1D model is designed with the help of basic block described above. The input signal 1D vector is required to be passed into a 1D convolution operation of kernel filter size of 7x7 with a sliding stride of 2 and padding of 3 so as to maintain the output channel size. After this, the batch normalization and activation ReLU is applied and output vector obtained is passed through a series of basic block consisting of two 3x3 layers as described above. Thus the series of such blocks becomes [3, 3, 3, 3] and making a total of 26 layered 1D model. The output so obtained is passed through a layer of pooling to avoid overfitting and 1D convolution model is converted into a linear fully connected dense network with SoftMax to predict the output classes. The architecture of the proposed model is shown in Figure 4.

6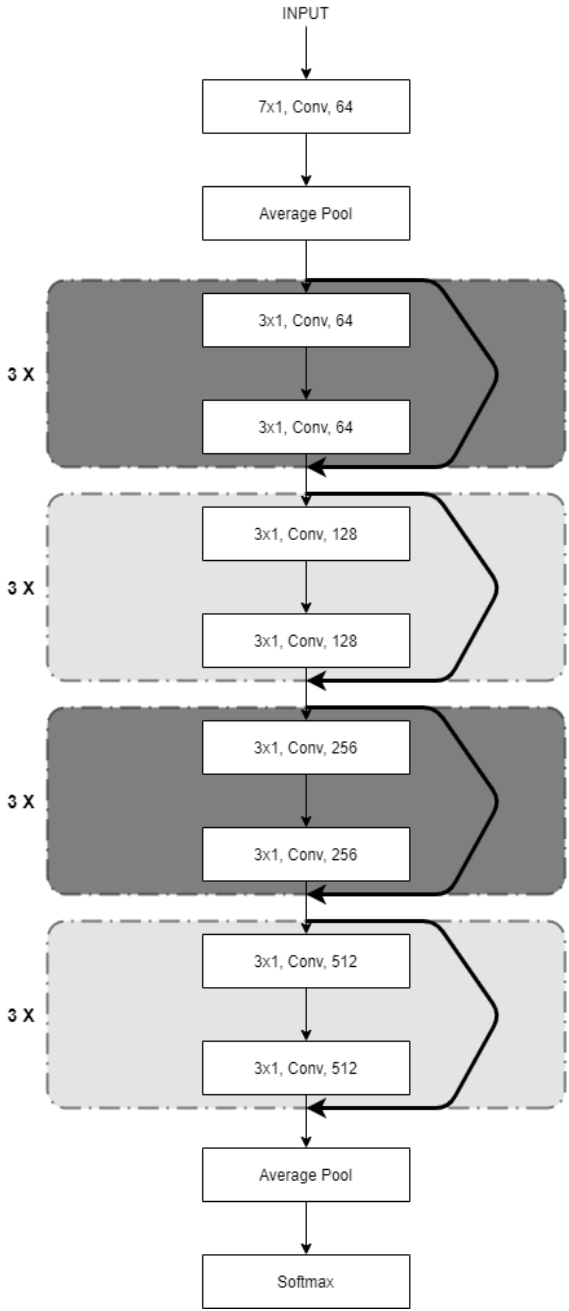

Figure 4. Proposed model

### 3.3 PROPOSED ALGORITHMS

**1. Algorithm for Basic Block with Input Vector V and Number of Neurons N.**

Step I:   Input data vector V into Basic Block
Step II:  Apply a convolution layer of 3x1 having stride value 1
Step III: Apply batch normalization
Step IV:  Apply Activation Function ReLU
Step V:   Repeat 2 and 3 for once
Step VI:  Add input vector V to step 5
          //Introducing skip connection to
          //solve vanishing gradient
Step VII: Apply ReLU again on step 6
Step VIII: Return the obtained featured vector map

**2. Algorithm for proposed model with Input EEG Signal vector E, and Number of Classes C**

Step I:    Input data vector E
Step II:   Apply a convolution layer of 7x1 having stride value 2 and padding of 3
Step III:  Apply batch normalization
Step IV:   Apply Activation Function ReLU
Step V:    Apply Max Pooling of filter size of 3x1, stride value 2, Padding value 1
           //reducing overfitting and
           //computational complexity
Step VI:   Repeat 3 times
           Call Basic Block Algorithm on step 5 with N value of 64
Step VII:  Repeat 3 times
           Call Basic Block Algorithm on step 5 with N value of 128
Step VIII: Repeat 3 times
           Call Basic Block Algorithm on step 5 with N value of 256
Step IX:   Repeat 3 times
           Call Basic Block Algorithm on step 5 with N value of 512
Step X:    Apply Average Pooling of stride value 1
Step XI:   Convert the neural network into fully connected network of 512 neurons
Step XII:  Reduce the number of neurons equal to number of classes C
Step XIII: Apply Softmax to find the probability of each classes
Step XIV:  Print the class having maximum probability  // Predicted Class

### 3.4 IMPLEMENTATION METHODOLOGY

The proposed model is coded with the help pytorch library available in python. Standard neural architectures such as LeNet , AlexNet

, VGG and DenseNet are executed on the dataset to find out the appropriate depth of network to be used in the proposed model. This will also help in the comparison of efficiency of proposed model on various Tasks.

A total of 20 epochs are used for training the models with 76% of total dataset as training dataset An epoch is basically refers to a complete traversal of whole dataset After each epoch the model is validated against next 12% of left out dataset and rest 12% of dataset is used for testing purpose. This is basically done so as to prevent overfitting of model while learning. Best model found during any epoch is being saved into external file storing the values of learned weights and hyper parameters which will be later used for testing and future use.

## 4. RESULTS AND DISCUSSION

In the proposed model as well as in the standard models, 20 epochs are used for training purpose. Training accuracy, Validation accuracy are measured for each model for all the Tasks viz. Task1, Task2, Task3, and Task4, as defined in the earlier section. 'Matplotlib' is used to plot the trends so observed in the model. Dataset used for the proposed model is also used in the training of standards models such as LeNet, AlexNet, DenseNet-121.

### 4.1 PROPOSED MODEL EXECUTION

**Task 1:** To identify the healthy brain activities (A, B) from brain during epilepsy seizure (E).

> In the training session of the proposed model for the Task 1, there is a sharp increase. This reflects that model is able to learn a lot of features. In validation session, it didn't performed so good in the starting epoch but over time after near to 20 epochs, the training accuracy and validation accuracy are very close to each other. This shows that the developed model is best for the problem. The observation are line plotted for the training and validation accuracy as shown in Figure 5.

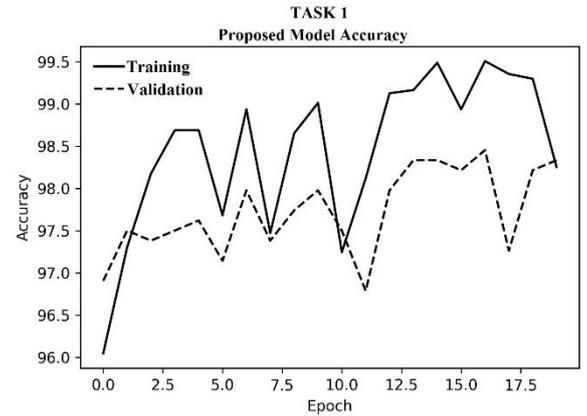

Figure 5. Training vs. Validation accuracy for task 1

Confusion matrix generated while performing Testing session is shown in Figure 6.

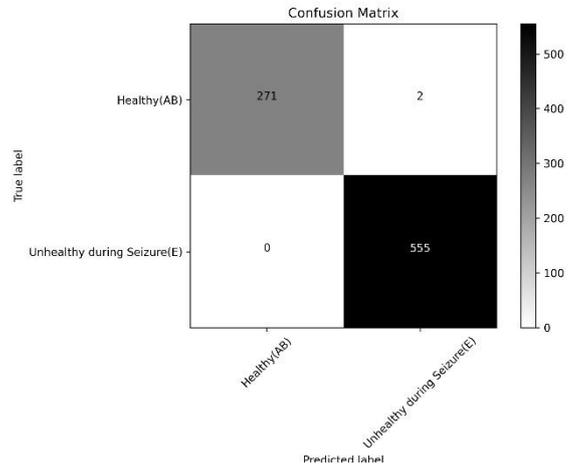

Figure 6. Confusion matrix for task 1

**Task 2:** To identify the healthy (A, B) wellbeing from epileptic patient (C, D, and E).

> In the execution of Task 2, Training accuracy follows the continuous increasing trend up to 20 epochs. The validation accuracy on the other hand follows a hilly trend due to fact that in start the learning of features is not quite effective so validation does not seem to work so effective but after some stages of learning by passing training set for around 8 epochs the validation accuracy increases but remains lower than training accuracy showing that model is not able to learn much of the features from the given data and model needs some more hyper tuning to perform better. Figure 7. shows the line plot of observation for the training and validation accuracy whereas Figure 8. shows the confusion matrix for the Testing accuracy.

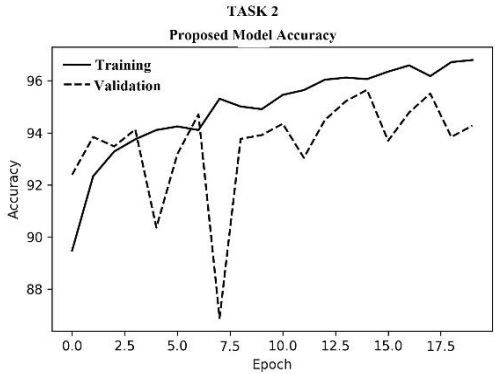

Figure 7. Training vs. Validation testing accuracy for task 2

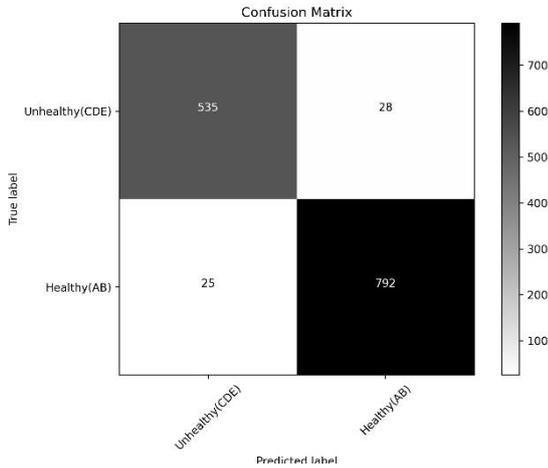

Figure 8. Confusion matrix for task 2

**Task 3:** For seizure identification between healthy (A, B), pre-ictal(C, D) and ictal (E).

For the Task 3 in the proposed model, the accuracy of the training process keeps on increasing up to 20 epochs and validation accuracy on the other hand is initially higher than the training accuracy. It makes model look like best model for assigned Task but there is a steep downfall near to 13$^{th}$ epoch due to sudden random weight updation but thereafter the performance of proposed model is again improved and remains very close to training accuracy. While reaching to 20th epoch, validation accuracy has higher slope than training accuracy as shown in the Figure 9. The confusion matrix shown in the Figure 10, is generated during Testing session.

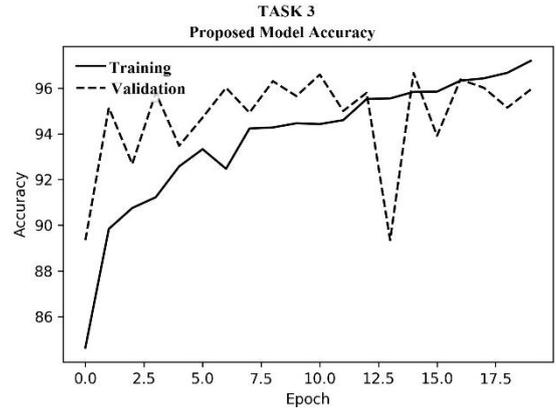

Figure 9. Training Vs. Validation testing accuracy for task 3

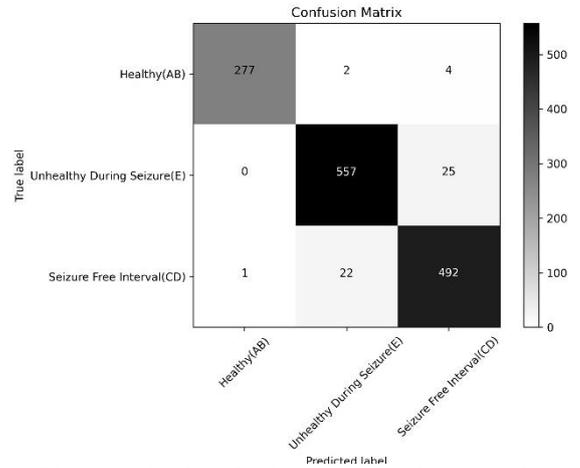

Figure 10. Confusion matrix for task 3

**Task 4:** To identify between all the five states i.e. Normal with open eyes (A), Normal with closed eyes (B), non-seizure unhealthy brain C and D and during seizure E. identify the healthy brain activities (A, B) from brain during epilepsy seizure (E).

Figure 11. depicts the line plot of Training accuracy vs. Validation accuracy. During the training session, the training accuracy of the proposed model increases exponentially up to 18 epochs. The validation accuracy on the other hand is higher than training accuracy initially but after certain epochs keeps increasing though not steeply and the testing performed after all the epochs and learning gives highest accuracy among all the 1D model in the Task 4. Confusion matrix generated during Testing session is shown in the Figure 12.

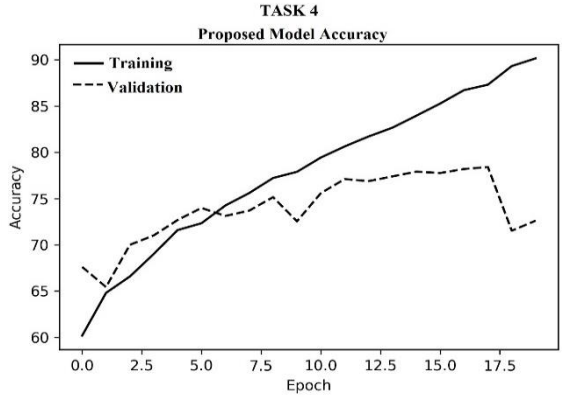

Figure 11. Training vs. Validation testing accuracy for task 4

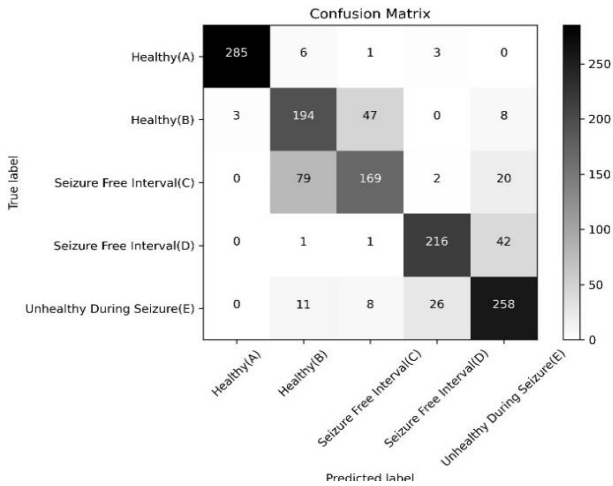

Figure 12. Confusion matrix for task 4

### 4.2 COMPARATIVE ANALYSIS OF PROPOSED MODEL WITH EXISTING ARCHITECHTURES

Table 2 to Table 5 compares the Training accuracy and Testing accuracy of proposed model with existing standard models. The existing standards models were also executed on the same dataset to have normalized accuracy for all the Tasks viz. Task1, Task2, Task3, and Task4.

Table 2. Training and testing accuracy Comparison For Task 1

| Architecture | Training Accuracy | Testing Accuracy |
|---|---|---|
| LeNet | 97.62% | 97.38% |
| AlexNet | 98.10% | 98.21% |
| DenseNet-121 | 98.33% | 97.62% |
| Proposed Model | **97.6%** | **98.33%** |

Table 3. Training and Testing accuracy comparison for task 2

| Architecture | Training Accuracy | Testing Accuracy |
|---|---|---|
| LeNet | 91.59% | 91.52% |
| AlexNet | 94.64% | 95.65% |
| DenseNet-121 | 95.80% | 95.58% |
| Proposed Model | **95.88%** | **96.16%** |

Table 4. Training and Testing accuracy comparison for task 3

| Architecture | Training Accuracy | Testing Accuracy |
|---|---|---|
| LeNet | 93.55% | 92.46% |
| AlexNet | 95.87% | 95.29% |
| DenseNet-121 | 97.54% | 94.71% |
| Proposed Model | **95.48%** | **96.81%** |

Table 5. Training and Testing accuracy comparison for task 4

| Architecture | Training Accuracy | Testing Accuracy |
|---|---|---|
| LeNet | 71.74% | 72.32% |
| AlexNet | 71.30% | 74.57% |
| DenseNet-121 | 79.82% | 76.74% |
| Proposed Model | **80.12%** | **81.30%** |

## 5. CONCLUSION

In the proposed model, 1D Convolutional Neural Network is used which overcomes the problem of vanishing gradient and also protects the true nature of input data (textual). In comparison to existing work, that are converting the data into two dimensional either in the form of matrix or converting the data into images and graphs.

It is concluded from the Tables 2 to 5 that the proposed model is better than the existing standard models in both Training and Testing accuracy for all the Tasks. In Task 1, the proposed model has 23% better precision than the best related present work. Further, the extraction of features which is much more complex in Task 4 there also proposed model outperformed the Training and Testing accuracy over the other standard models LeNet, AlexNet, DenseNet-121 as well as when compared to present related work carried out in Task 4. There is significant increase of 7% accuracy while using 1D model instead of 2D models used by other researchers.

**DISCLOSURE STATEMENT**

No potential conflict of interest was reported by the authors.